# A Framework For Intelligent Multi Agent System Based Neural Network Classification Model

Roya Asadi[1], Norwati Mustapha[2], Nasir Sulaiman[3]

[1,2,3]Faculty of Computer Science and Information Technology,

University Putra Malaysia, 43400 Serdang, Selangor, Malaysia.

[1]royaasadi@yahoo.com, [2,3]{norwati, nasir}@fsktm.upm.edu.my)

*Abstract*—**Intelligent multi agent systems have great potentials to use in different purposes and research areas. One of the important issues to apply intelligent multi agent systems in real world and virtual environment is to develop a framework that support machine learning model to reflect the whole complexity of the real world. In this paper, we proposed a framework of intelligent agent based neural network classification model to solve the problem of gap between two applicable flows of intelligent multi agent technology and learning model from real environment. We consider the new Supervised Multi-layers Feed Forward Neural Network (SMFFNN) model as an intelligent classification for learning model in the framework. The framework earns the information from the respective environment and its behavior can be recognized by the weights. Therefore, the SMFFNN model that lies in the framework will give more benefits in finding the suitable information and the real weights from the environment which result for better recognition. The framework is applicable to different domains successfully and for the potential case study, the clinical organization and its domain is considered for the proposed framework.**

## I. INTRODUCTION

Neural networks are the branch of psychology and neurobiology for developing and testing computational analogues of neurons. The best features in neural networks are their non-missing values and high tolerance to noisy data, as well as their ability to classify data pattern on which they have not been trained. The main advantage of neural networks lies in its scaling and learning abilities as intelligent system in machine learning algorithms [4].

Multi agent technology is applied by intelligent systems to solve the problems of analysis of complex systems and intelligent management activities. Intelligent Multi Agent Systems (MAS) based learning combine collection of information from their environment, recognition data, intelligent classification data and prediction future data, storage data, delivery data to knowledge management systems such as Decision Support System (DSS) and Management Information System (MIS) [1, 5, 2].

Currently, there is the lack of one united framework for combination of the two applicable flows of intelligent multi agent technology and learning in real environment [15, 1]. For solving this gap, we consider new SMFFNN model as intelligent core of intelligent agent based learning framework [13]. The framework earns the information from the respective environment and its behavior can be recognized by the weights.

Supervised Multi-layer Neural Network (SMNN) models need suitable data pre-processing techniques to find input values while pre-training techniques to find desirable weights that in turn will reduce the training process. Without preprocessing, the classification process will be very slow and it may not even complete. Potential Weights Linear Analysis (PWLA) is new technique for reducing training process and fast classification in new SMFFNN model with high accuracy. The first PWLA normalizes input values as data preprocessing and then uses normalized values for pre-training, at last reduces dimension of normalized input values by using their potential weights. SMNN models can changed to new models by using PWLA [12]. All agents of system can apply the outputs of PWLA technique and new SMFFNN model.

This paper is organized as follows: To discuss and survey some related works in intelligent agents based systems and their components. Proposed framework is intelligent agent system based neural network classification by using new SMFFNN model and PWLA technique. Clinical organization and its domain are considered for illustration of our framework. Finally, the conclusion with future works is reported respectively.

## II. RELATED WORKS

In this section, we discuss two applicable flows of intelligent agent based system and neural network model as learning model. In neural network section, we consider new supervised multi-layer feed forward neural network model and PWLA as preprocessing and pre-training technique exclusively in using our proposed methodology.

### A. Intelligent Multi Agent Based System

Intelligent is general terms to describe some of mind activities such as judging, logical thinking, planning and learning. Artificial Intelligence (AI) is based on intelligent behavior. Machine learning, expert systems and data mining

---

such as neural network models can help to implement AI for learning of virtual environments. Intelligent Agent (IA) is made by using artificial intelligence properties [3, 14, 10]. Wooldridge (2002) and Langseth et al. (2005) explained some aspects of IA which are capable of flexible autonomous action to meet their design objectives [17, 5]. They are:

- Reactivity: IA receives information of its environment by its sensors, changes internal design objectives of its structure and has suitable actions with feedback periodically.

- Pro–activeness: IA can show goal directed behavior by taking the initiative, responding to changes in their environment in order to satisfy their design objectives.

- Sociability: IA has capability of interacting with other agents for negotiation and/or cooperation to satisfy their design objectives.

Other properties of IA are self-analysis, learning, adapting and improving through interaction with the environment.

Padghan and Winikopff (2004) explained that the term of Agent refers to an entity that acts on behalf of other entities or organizations; and Multi Agent System (MAS) consists of several agents with capable of common interaction with self-organization [11]. Generally, the structure of multi agent system is as follow:

- Actions: Respons.ing of agent in front of environment events and changes,

- Percepts: Accumulating information from the environment,

- Events: Processing of updating beliefs and to operate actions,

- Goals: Considering objectives of system to accomplish and can be updated,

- Beliefs: Handling accumulated information about the environment,

- Plans: Using plan library for handling events and achieve goals,

- Messages: Necessary for agents to interact,

- Protocols: Rules of interaction.

Yoav and Kevin (2009) and Michael (2001) explained about Agent Based System (ABS) and intelligent agent as two relevant streams and the gap of one united framework for combination these two [18, 9]. Intelligent multi agent systems have great potentials and relevant to use in different research areas especially in virtual environments to support machine learning model with the whole complexity of the real world [15].

Bobek and Perko (2006) showed that intelligent agents can be used for [1]:

- Intelligent Acquisition: To collect unstructured data from environment of system. The suitable input values and their weights will be recognized by multi-agents.

- Intelligent modeling: To create intelligent agent based system frameworks in predicting of data handling, future events and intelligent rules and plans.

- Intelligent delivery: To proactive delivery of selected the suitable information and advanced report to approach the special strategies.

### B. Neural networks and Learning system

Neural networks are suitable for extracting rules, quantitative evaluation of these rules, clustering, self-organization, classification, regression feature evaluation, and dimensionality reduction. Learning is the important property of neural networks. Neural networks are able to dynamically learn types of input information based on their weights and properties. Suitable data preprocessing and pre-training techniques are necessary to find input values while pre-training techniques to find desirable weights that in turn will reduce the training process. This is the essence of Supervised Multi-layer Neural Network (SMNN) [4, 8] that will be used in our framework.

Perspectives of learning consist of three majors: supervised learning, unsupervised learning, and reinforcement learning. Supervised training is similar to unsupervised training in the sense that training sets are provided. The difference between these two is that in supervised training, the desired output is provided and weight matrix is applied based on the difference between the predicted output and the actual output of the neural network. In reinforcement learning, the model has capable of generating certain effects, interactions to the environment and estimation of the unknown item. Neural networks are frequently used in reinforcement learning as part of the overall algorithm. The tasks include control problems and sequential decision making tasks [6, 8].

The best features in neural networks are their non-missing values and high tolerance to noisy data, as well as their ability to classify data patterns on which they have not been trained. The best and popular method in neural networks is back-propagation network (BPN) by Werbos (1974) as Supervised Multi-layer Neural Network model [4, 16].

Back-propagation network learning uses gradient-based optimization methods in two basic steps: calculate the gradient of error function, and employ the gradient. The optimization procedure includes of a high number of small steps, causing the learning to be considerably slow [7]. Optimization problem in supervised learning can be shown as sum of squared errors between output activations and target activations in neural network as well as the minimum weights. BPN can be changed to new SMFFNN model with high accuracy and low processing time by using PWLA technique [12, 13].

#### 1) Potential Weights Linear Analysis

Potential Weights Linear Analysis (PWLA) is one technique for reducing training process and fast classification in new SMFFNN model with high accuracy [12]. The key ideas in PWLA technique and new SMFFNN model are to recognize high deviations of input values matrix from global mean and next is pre-training using the meaning of vector torque formula. The aim of applying PWLA technique is to recognize and find which input values have more differences and deviations to global mean. These deviations cause more scores for their values. After illustration of weak and strong weights, PWLA can omit weak weights and just enter strong weights to training process of SMFFNN model. Input values of PWLA can be from numeric type, range and measurement unit. If the dataset is large, there is a high chance that the vector components are highly correlated (redundant). PWLA method is able to solve this problem. There are three phases in implementing PWLA:

*a) Data preprocessing:* This phase is normalization of input values. The technique of "Min and Max" is used. Each value is computed to find the ratio to average of all columns.

*b) Pre-training:* In improving pre-training performance, potential weights are initialized. The first distribution of standard normalized values is computed. Therefore, PWLA does not need to use any random number for initialization of potential real weights and input of pre-training phase is normalized input values. The input values vectors create vector torques. Distributions of standard normalized values show the arms of vector torques which are weights. Global mean is the center of vectors torques. The weights show deviations of input values matrix from global mean. Each weight is equivalent to sum of all absolute of normal values in each instance. This definition of weight is based on statistical and mathematical definition of normalization distribution and vector torque.

*c) Dimension reduction:* PWLA can map high dimension matrix to lower dimension matrix based on strong potential weights, and the suitable sub-matrix of necessary attributes can be selected. The strong weight causes high variance. If the dimension of the input vectors be large, the components of the vectors are highly correlated (redundant). PWLA can solve this problem in two ways. First, after equivalence, the global mean take place as one constant at special point of axis of vector torques and the weights are distributed between vectors. In other way, it can solve redundancy by dimension reduction. This phase of PWLA can be performed in hidden layer during pruning. PWLA technique outputs are dimension reduction of normalized input values and potential weights.

*2) New Supervised Multi-layer Feed Forward Neural Network Model (SMFFNN)*

New SMFFNN model will process based on the algebraic consequence of vectors torques [13]. Each torque shows a real worth of each value between whole values in matrix. Recall that BPN uses sigmoid activation function to transform actual output between domain [0, 1] and to compute error by using the derivative of the logistic function in order to compare actual output with true output. True output forms the basis of the class labels in a given training dataset.

Here, PWLA computes potential weights and new SMFFNN computes desired output by using binary step function instead of sigmoid function as activation function. Also there is no need to compute error and the derivative of the logistic function for the purpose of comparison between the actual output and the true output. New SMFFNN applies

PWLA similar to the simple neural network. The number of layers, nodes, weights, and thresholds in new SMFFNN using PWLA pre-processing is logically clear without presence of any random elements. New SMFFNN will classify input data by using output of PWLA, whereby there is one input layer with several input nodes, one hidden layer, and one output layer with one node. Here, new SMFFNN exists only in one epoch during training processing without the need to compute bias and error in hidden layer. In evaluating the test set and predicting the class label, weights and thresholds are clear and class label of each instance can be predicted by binary step function.

### III) A FRAMEWORK FOR INTELLIGENT MULTI AGENT SYSTEM BASED NEURAL NETWORK CLASSIFICATION MODEL

This section has a discussion on solving the gap of lacking in one united framework for combination of two relevant flows which are intelligent multi agent systems in real world and learning systems. The proposed framework in virtual environment is intelligent agent based neural network classification using new SMFFNN model. For illustration of framework, we select the clinical organization and its environment. There are several issues in this system such as intelligent acquisition, intelligent modeling and intelligent delivery. Figure 1 shows outline of intelligent multi agent based neural network classification system.

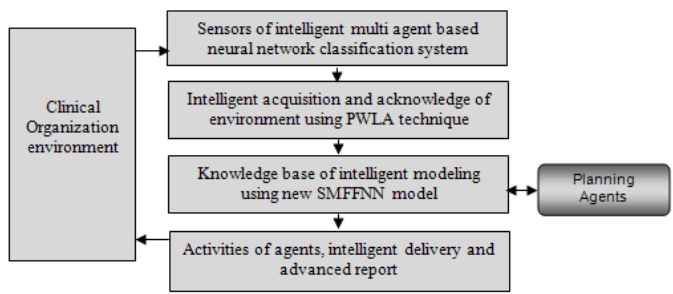

Figure 1.    Outline of intelligent multi agent based neural network classification system

• Intelligent acquisition: To generate knowledge based on gathering data by PWLA preprocessing and pre-training technique about population, health centers, distribution of offices and branches of clinical organization, staffs, etc. Outputs of PWLA are normalized input values and potential real weights.

• Intelligent modeling: To handle data, rules, plans, and future prediction. In this paper, one framework using new SMFFNN model for intelligent classification with high accuracy and speed is proposed [13]. This framework earns its information from the respective environment and PWLA technique helps to prepare suitable and real values and weights. New SMFFNN model uses outputs of PWLA and classifies and predicts desired output. If selection of the attributes and input values be good and suitable, the accuracy of classification can be reach to 100% [12]. The intelligent multi agent framework has several agents that apply the outputs of PWLA technique and the output of new SMFFNN model. The argument is capable of handling regulated, structured data, enhanced with unstructured data, derived from various and distributed sources, investment records, plans, realizations and etc.

• Intelligent delivery: Advanced reporting based on managers' views by using management and decision systems such as MIS and DSS, and so ability to proactively respond to exceptional events, to decide and upgrade rules and plans. Output of framework will be intelligent results and advanced report via MIS, DSS and so on that they help managers to act best rules in front of events of environment. This intelligent agent system has feedback for updating believes and rules too.

Suppose that the government tries to cover all over the country with health network. If it is considered one branch of clinical organization with full facilities and appliances for each zone, it is not economical and it has a lot of cost. Therefore the government's analyzers and designers have to classify the zones based on two class labels. The zones with class label 1 are main or central zones and the zones with class label 2 are depended zones. The government constructs or improves big hospital with full facilities and appliances for main zones. Main zones must support depended zones. So depended zones will have health centers or clinics with limited facilities. Classification of zones and management of clinical organization are very difficult in the traditional system and using an intelligent classification model is necessary. Therefore, a clinical organization based on multi agent system must be considered and then an intelligent agent for classification with high accuracy and low processing time must be considered as well.

The traditional organization chart has several problems in managing, especially in managing of information and earning the desired outputs such as redundancy of data, multiple updating in Data Base Management System (DBMS). The traditional chart can be changed to intelligent multi agent system based neural network classification.

The figure 2 shows pyramid of clinical organization which is one simple purposed chart with two main agents and one head.

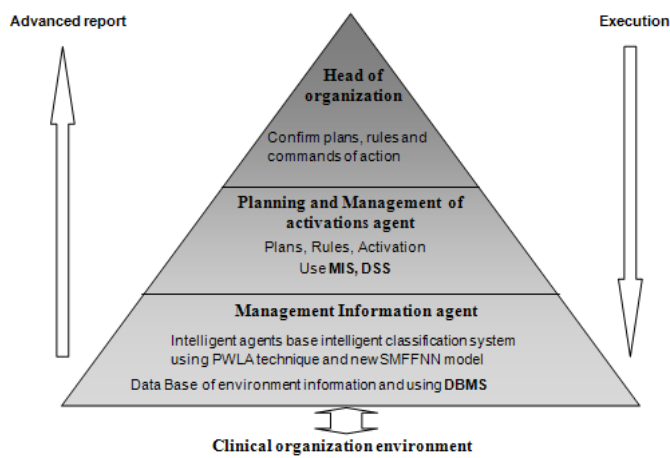

Figure 2. Pyramid of clinical organization

The management information agent collects the information from Clinical environment. Usually, the information is distributed and we need one system with several branches for gathering data. Therefore, there are several different sub-agents in this system such as agent of intelligent classification, agent of staff management, agent of facilities management and etc.

Figure 3 shows details of the management information agent of clinical organization:

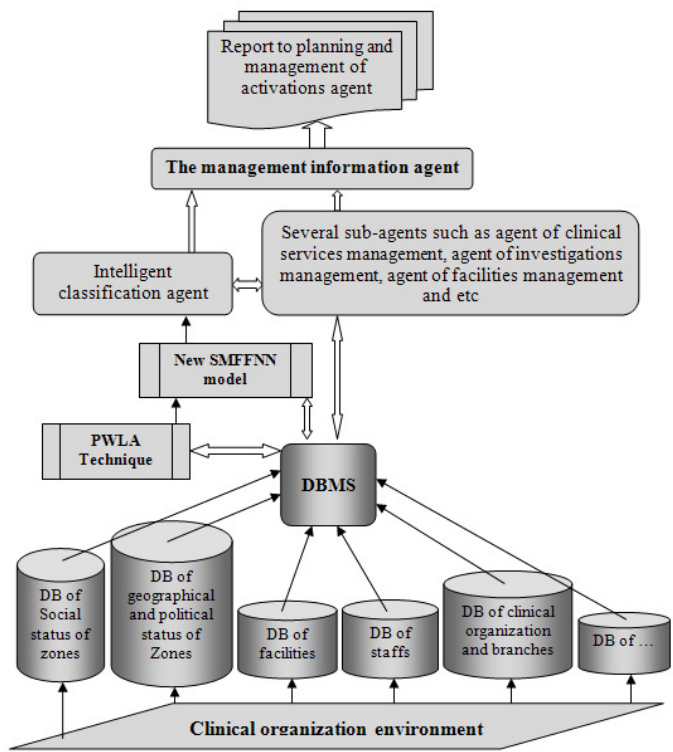

Figure 3. Details of the management information agent of clinical organization

The information is collected in one database and classified by intelligent classification model. The information of clinical environment is stored in one database, and it will be controlled in the Data Base Management System by DBA (Data Base Administrator).

Advantages of DBMS are:

• Minimizing redundancy, high security, integrity and reliability of data

• Considering three layers for data: Internal/physical layer, Conceptual/abstract layer, External layer/high level view for query.

• Creating relational dataset

DBMS collects different databases from clinical environment with their attributes, instances and relation sets. Intelligent classification agent classifies instances and predicts their situations, based on different goals of system and selected attributes, by sub-agents of management information agent.

The management information agent sends the reports of knowledge to planning and management of activation agent for applying by DSS, MIS, statistical and scientific software, and other management systems. The plans and rules will be updated and advanced reports will send to organization head for confirmation and action permission.

For best clinical services to all people in all over the country, the country must be considered as environment of system. Health centers, hospitals and branches of clinical organization must be distributed on everywhere of the country. For approach to this goal, we select one type of two special clinical organization chart types for every zone based on social, political and economical conditions of the zones. Therefore, the zones must be classified based on their information. Predicting the class label of each zone will help to managers which type of charts is suitable for each zone. Figures 4 and 5 show two special clinical organization chart types of A and B [http://www.tpub.com/content/armymedical/md0750/md07500017.htm] as follow:

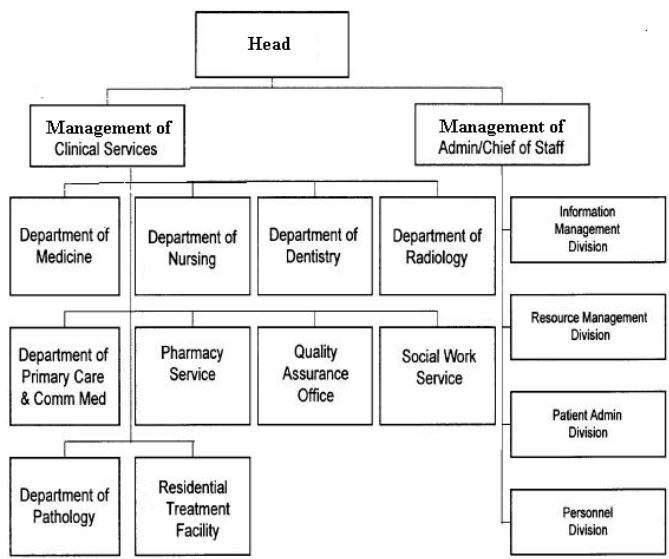

Figure 5. Type B: Health center with simple organization for small and depended zones

Main zones will have big hospitals with complex organization and they will be selected as central zones for depended zones in around of them. Small zones are depended zones and will have health centers with simple organization. The managers must select necessary attributes to classify zones and PWLA technique is able to help them. They must illustrate which zones must be centers, which zones must be around of central zones and use conveniences and facilities of them. For classification of zones, the first necessary attributes of each database must be selected. For example, they can select eight attributes: 1-City population of each zone, Rural population of each zone from database of Social status of zones, 2-Area of each zone, 3-The number of neighbors of each zone, 4-The distance between each zone and capital from database of 5-Geographical and Political status of zones, 6-The number of local employees in each zone from database of Staffs in each zone, 7-The number of persons with medical insurance in each zone, 8-The number of health centers or hospital in each zone from database of Clinical organization and its branches in each zones. Subagent of intelligent classification receives those attributes via DBMS in physical layer and classified zones by using new SMFFNN model and PWLA. Figure 6 shows intelligent classification of zones.

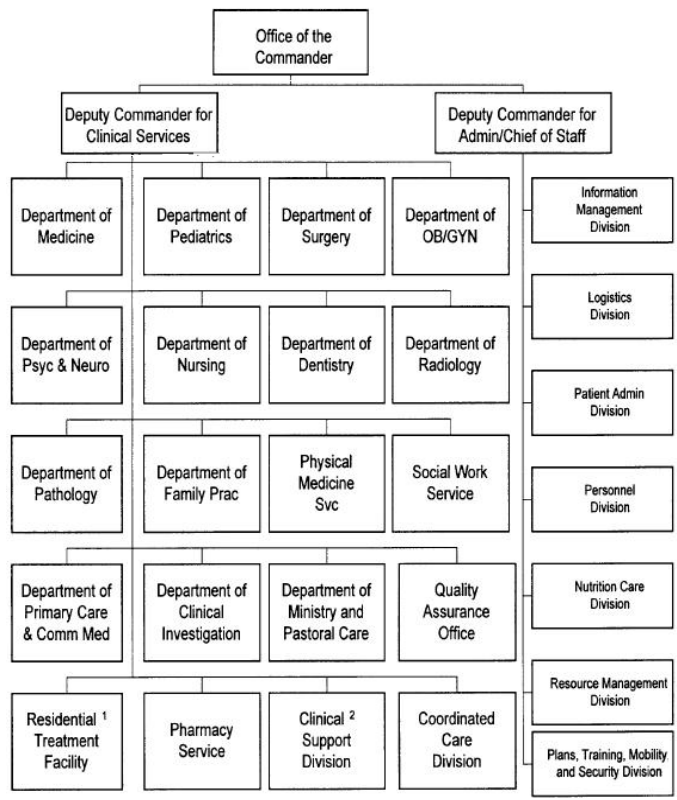

Figure 4.  Type A: Big hospital with complex organization for central or main zones

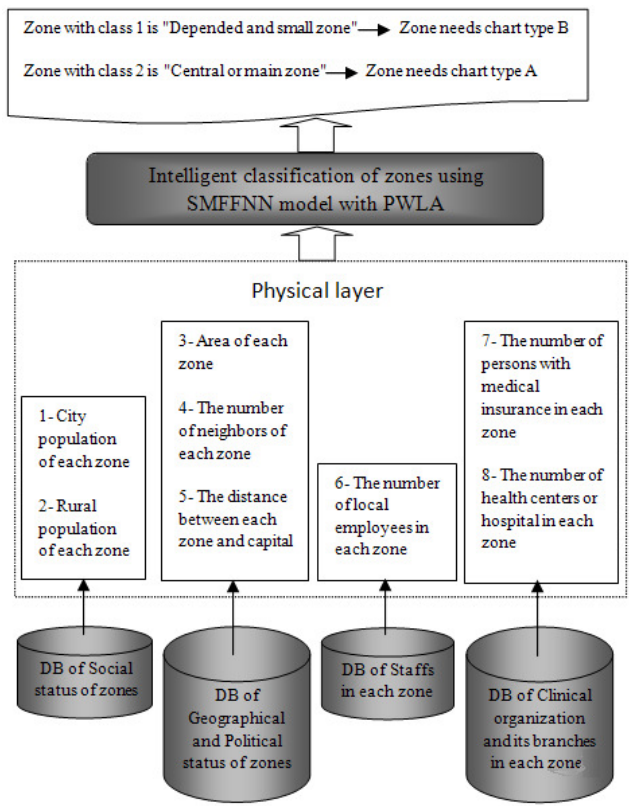

Figure 6. Intelligent classification of zones

Physical layer earns its input values of necessary attributes for classification of zones from different data bases of system. The class of each zone will show that the zone is main zone or depended zone and which type of clinical organization charts is necessary for this zone. Figure 7 shows four main zones with big equipped hospital that are able to cover and support depended zones surround them.

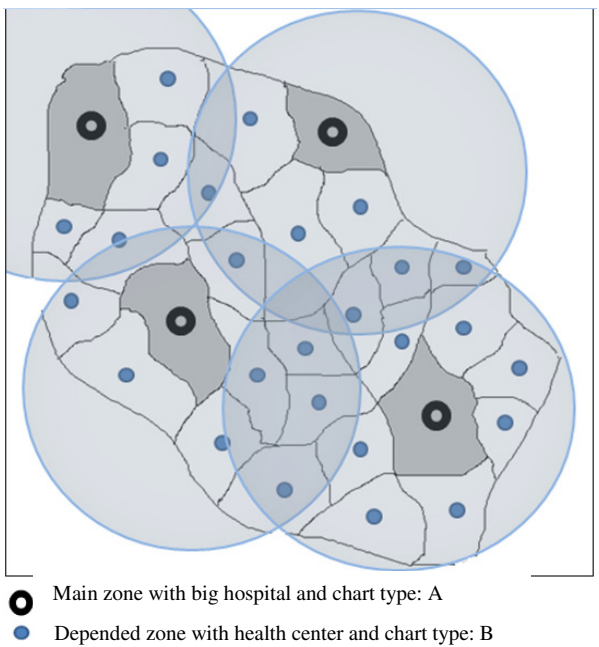

○ Main zone with big hospital and chart type: A

● Depended zone with health center and chart type: B

Figure 7. Covering all depended zones by four main central zones

This example was one case which applies intelligent classification. There are several goals and plans in this system that they need intelligent classification. These processes can be run in parallel. Intelligent multi agent based neural network classification system has capability of predicting future status and showing complex relations between components of systems.

## IV) CONCLUSIONS AND FUTURE WORK

Agent based system and intelligent agents are two important issues in research area. Intelligent multi agent systems have great potentials to use in different purposes. We consider the problem of lacking in one united framework for combination of two relevant flows which are intelligent multi agent systems in real world and learning systems for creation one intelligent framework in virtual environment. The traditional prototypes and old systems can be changed to intelligent multi agent system with clear components, relationships, and better control of them. We discussed about this framework by using new SMFFNN model as neural network classification in the environment of the clinical organization. As mentioned that the new SMFFNN model is derived by applying PWLA technique to BPN. Therefore, other neural network models can also be changed to the new one by applying PWLA as well. Then, for future work, we will improve this intelligent framework by using other new models and apply them to other environments.

### Aᴜᴛʜᴏʀs ᴘʀᴏꜰɪʟᴇ

**Dr. Prof. Md. Nasir bin Sulaiman** is a lecturer in Computer
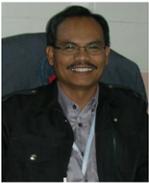
Science in Faculty of Computer Science and Information Technology, UPM and as an Associate Professor since 2002. He obtained Ph. D in Neural Network Simulation from Loughborough University, U.K. in 1994. His research interests include intelligent computing, software agents and data mining.

**Dr. Norwati Mustapha** is a lecturer in Computer Science in
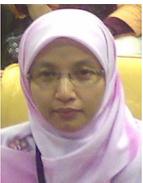
Faculty of Computer Science and Information Technology, UPM and head of department of Computer Science since 2005. She obtained Ph. D in Artificial Intelligence from UPM, Malaysia in 2005.

Her research interests include intelligent computing and data mining.

**Roya Asadi** received the Bachelor degree in Computer
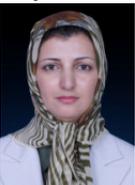
Software engineering from Electronical and Computer Engineering Faculty, Shahid Beheshti University and Computer Faculty of Data Processing Iran Co. (IBM), Tehran, Iran.

She is a research student of Master of Computer science in database systems in UPM university of Malaysia. Her professional working experience includes 12 years of service as Senior Planning Expert 1. Her interests are in Intelligent Systems and Neural Network modeling.